# MEMS Gyroscope Multi-Feature Calibration Using Machine Learning Technique

Yaoyao Long, Zhenming Liu, Cong Hao, Member, IEEE, Farrokh Ayazi, Fellow, IEEE

***Abstract*— Gyroscopes are crucial for accurate angular velocity measurements in navigation, stabilization, and control systems. MEMS gyroscopes offer advantages like compact size and low cost but suffer from errors and inaccuracies that are complex and time varying. This study leverages machine learning (ML) and uses multiple signals of the MEMS resonator gyroscope to improve its calibration. XGBoost, known for its high predictive accuracy and ability to handle complex, non-linear relationships, and MLP, recognized for its capability to model intricate patterns through multiple layers and hidden dimensions, are employed to enhance the calibration process. Our findings show that both XGBoost and MLP models significantly reduce noise and enhance accuracy and stability, outperforming the traditional calibration techniques. Despite higher computational costs, DL models are ideal for high-stakes applications, while ML models are efficient for consumer electronics and environmental monitoring. Both ML and DL models demonstrate the potential of advanced calibration techniques in enhancing MEMS gyroscope performance and calibration efficiency.**

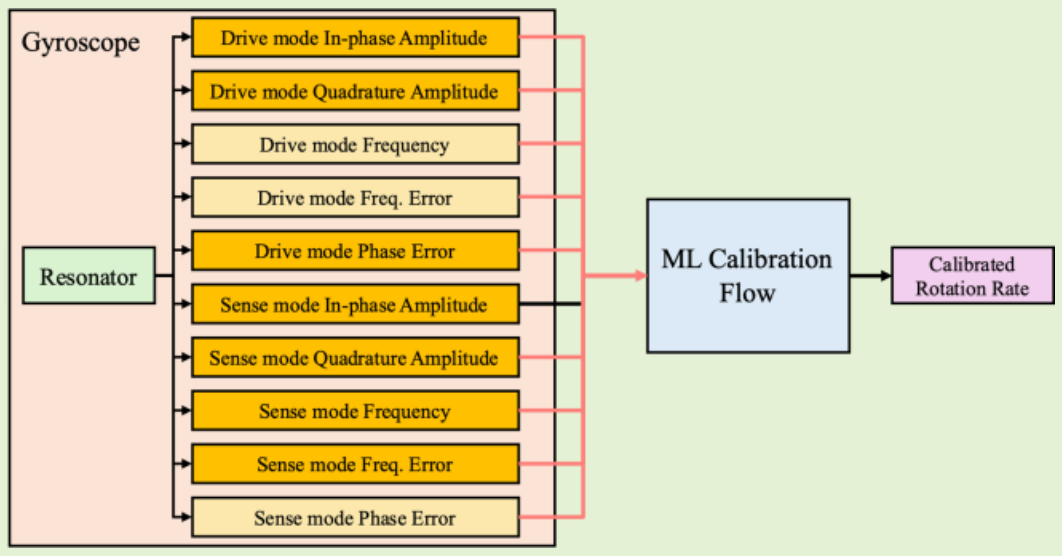

## I. Introduction

GYROSCOPES have a wide range of applications that require precision measurements of angular velocity in navigation, stabilization, automotive, drones, and consumer electronics. MEMS (Micro-Electro-Mechanical Systems) gyroscopes, in particular, offer significant advantages over traditional mechanical and optical gyroscopes due to their small size, low cost, and ability to integrate with electronic systems. These compact devices are extensively used in smartphones, gaming devices, drones, and automotive systems for their precise motion-sensing capabilities [1-3]. Nevertheless, most of the existing MEMS gyroscopes still struggle with significant errors such as bias instability, scale factor errors, and random drifts, which will degrade their performance [1-3]. Meanwhile, the traditional linear calibration methods, employed to derive a constant scale factor for each gyroscope, fail to capture the complex, non-linear relationships between the output signal and rotation rate, leading to reduced accuracy. Additionally, the calibration requires a controlled rate test for each device individually to extract the scale factor. This process is both time-consuming and costly due to the extensive manual labor and precise control needed for each test. Addressing these challenges necessitates the development of new calibration techniques that can mitigate the impact of noise, enhance the accuracy and reliability of MEMS gyroscopes, and eliminate the need for repetitive rate tests for identically designed gyroscopes.

Machine learning (ML) has emerged as a powerful tool in various technological domains thanks to its ability to learn from vast amounts of data, model non-linear characteristics, and improve prediction accuracy. Such applications include health condition predictions [4-6], weather and earthquake predictions [7, 8], finance management [9], and in the MEMS area to optimize device design [10], modeling [11], and compensate for temperature variation [12, 13]. One of the key strengths of ML

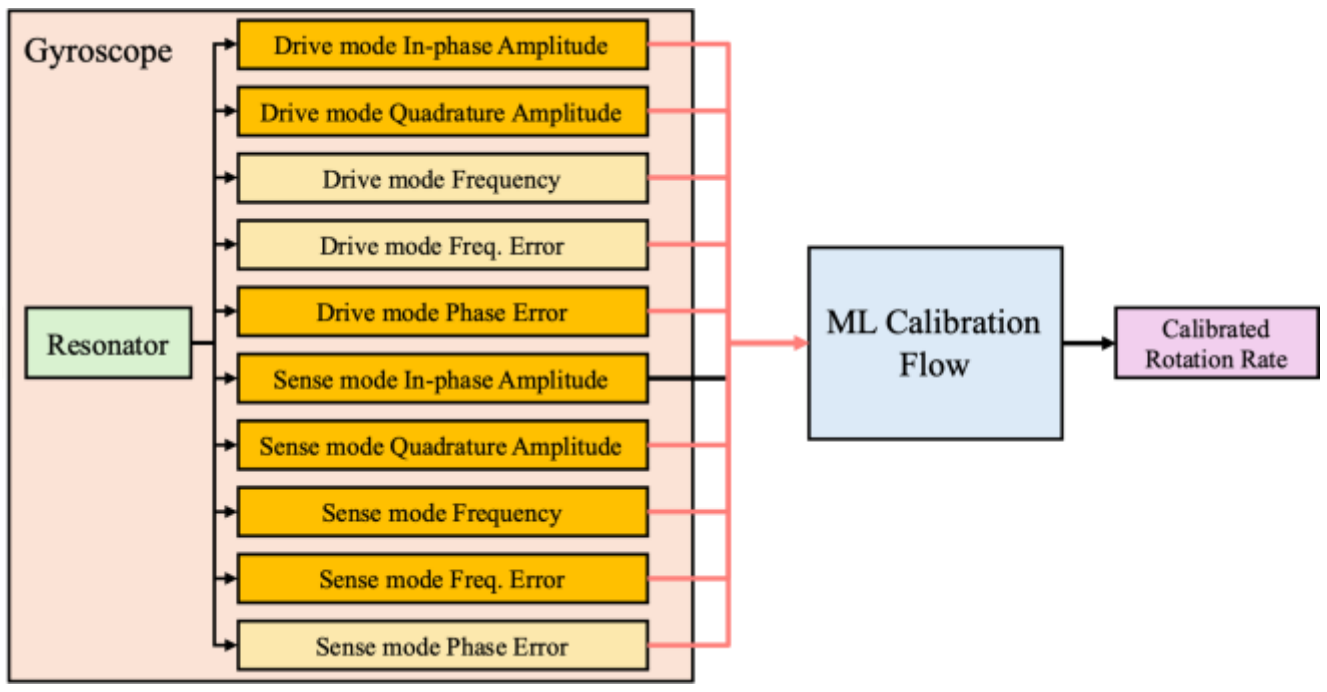

Fig. 1. The process flow of using ML and DL to directly calibrate MEMS gyroscope based on the multiple output signals/features of the resonator element. After feature analysis, the selected features are highlighted.

Yaoyao Long is with the School of Electrical Engineering and Computer Science, Georgia Institute of Technology, Atlanta, GA 30332, USA (e-mail: ylong60@gatech.edu).
Zhenming Liu is with StethX Microsystems, Atlanta, GA, 30308, USA (e-mail: zhenming@stethx.com).
Cong Hao is with the School of Electrical Engineering and Computer Science, Georgia Institute of Technology, Atlanta, GA 30332, USA (e-mail: callie.hao@gatech.edu)
Farrokh Ayazi is with the School of Electrical Engineering and Computer Science, Georgia Institute of Technology, Atlanta, GA 30332, USA (e-mail: ayazi@gatech.edu)

is its capacity to adapt to new data, making it suitable for applications requiring continuous learning and adaptation. Moreover, ML algorithms can also automate feature extraction, reducing the need for manual intervention and enabling the discovery of hidden patterns that traditional methods might miss.

Previous research has explored using ML for temperature compensation and steady-state denoising of MEMS gyroscopes [14-16]. However, these studies focused on stationary conditions, leaving issues like scale factor non-linearity and long-term instability in real-world applications unaddressed. By using a high-performance gyroscope as a reference, we can develop a model suitable for dynamic conditions. Notably, existing studies have primarily used the rate output, i.e. the sense mode in-phase signal, as the sole input feature to their ML models, neglecting other important features as listed in Fig. 1. Incorporating these features is crucial, as they could provide valuable information about the state of the resonant modes. For instance, the device resonant frequency is able to indicate environment temperature, which alters the scale factor in MEMS gyroscopes [17]; phase error in PLL can reflect signal frequency error [18]; and the quadrature signal signifies bias instability [19]. Therefore, the ability to handle complex and high-dimensional data of ML can offer significant advantages in the MEMS gyroscope calibration area, enhancing performance, reliability, and efficiency.

## II. Research Foundations

### A. AlN-on-Si Piezoelectric Resonant Gyroscope

In our previous work [20], we demonstrated an AlN-on-Si bulk acoustic wave (BAW) resonant gyroscope (Fig. 2), which exhibited state-of-the-art noise specifications when operated in vacuum. Specifically, it achieved an angle random walk (ARW) of 0.14°/√h and a bias instability (BI) below 10°/h. However, the optimal performance of this gyroscope heavily depends on mode-matching operations, where the natural frequencies of the drive and sense modes are equal, and the sense mode remains silent in the absence of rotation. Due to fabrication imperfections, mode-matching is not guaranteed. In [20], we employed laser trimming alongside a novel quadrature cancellation technique to achieve mode-matching in the as-fabricated devices. Additionally, the gyroscope needs to operate in a vacuum to attain a high Q-factor (10,000) in both resonant modes. However, in-air operation degrades its Q-factor and thus signal-to-noise ratio.

Fig. 3 illustrates the measured Allan deviation (ADEV) of unmatched AlN-on-Si gyroscopes when operated untrimmed and in-air, compared to a commercial MEMS gyroscope from the high-performance DMU11 inertial measurement unit (IMU) by Silicon Sensing [21]. The two AlN-on-Si gyroscopes used in this work operate at frequency of 3MHz and have in-air Q factors of 2,500 with gyroscopic mode splits of 320 to 440 Hz. The measured ADEV of these two gyroscopes shows a much inferior short-term and long-term performance compared to DMU11, which exhibits an ARW of 0.35°/√h and a BI of 3.53°/h. The significant difference between the performance of the resonant piezo gyro and DMU11 indicates the potential for using a higher-performing sensor to calibrate and denoise the lower-performing untrimmed piezo gyros operating in the air.

### B. Experimental Setup

The reference IMU is directly connected to a computer using its built-in application, with the output timestamp used to synchronize with the piezoelectric MEMS gyroscope and the output Z-axis rotation rate serving as the target for each time point. The piezoelectric MEMS gyroscopes in this study were operated using multi-coefficient eigenmode and measured with the HF2LI 50 MHz lock-in amplifier from Zurich Instruments. The output signals included timestamp, sense mode and drive mode in-phase and quadrature demodulated electric voltage signals, sense mode and drive mode resonant frequencies, sense mode and drive mode resonant frequency error, and sense mode and drive mode resonant phase error, as shown in Fig. 1. Among these signals, the sense mode demodulated in-phase signal is the primary signal, traditionally used to linearly fit a scale factor to derive the rotation rate.

In this research, the reference IMU and the piezoelectric MEMS gyroscopes are mounted on the same circuit board,

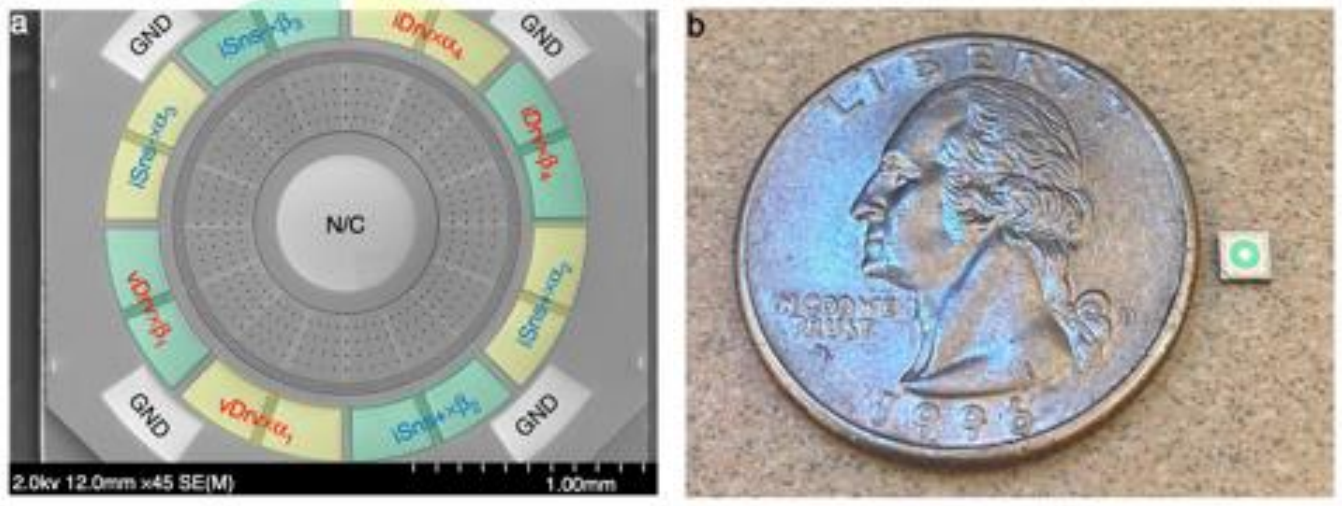

Fig. 2. (a) SEM image of a fabricated AlN-on-Si annulus resonator (b) 2.5 mm × 2.5 mm AlN gyroscope die was placed next to a US quarter for size comparison [20].

which is secured to a solid frame and affixed to the rate table with screws inside a temperature-controlled chamber. This configuration ensures that both the reference IMU and the piezoelectric MEMS gyroscopes measure the same Z-axis rotation rate simultaneously.

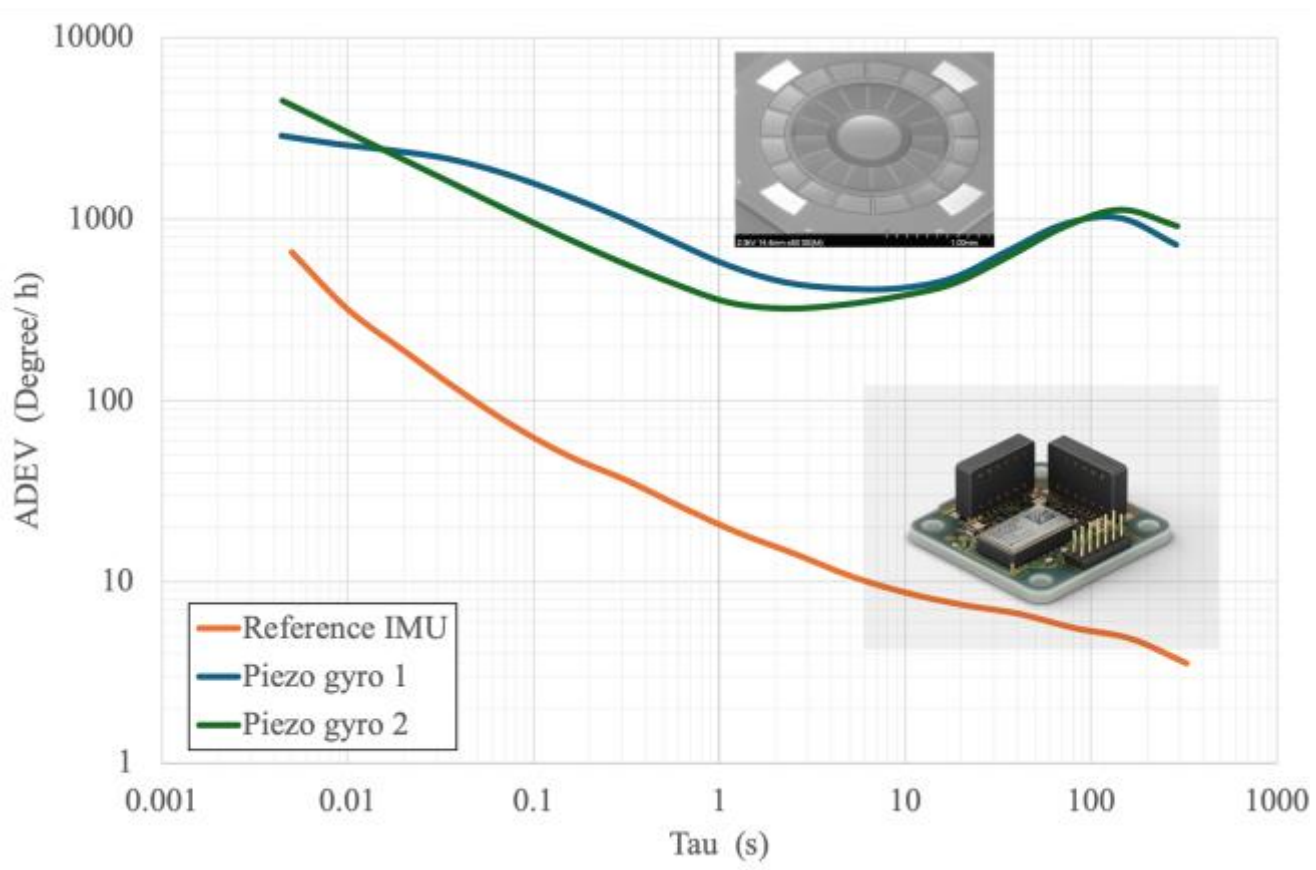

Fig. 3. ADEV plot and pictures of the AlN-on-Si piezoelectric MEMS gyroscopes and the reference IMU at room temperature in ambient air.

To mimic real application environments, the measurement process for both the training and testing process involved subjecting the rate table to random rotation rates and periods of steady state, where the rate table was locked at zero rotation rate. The training process was done with one of the AlN-on-Si gyroscopes, gyro 1. The test datasets were collected from both gyroscopes, gyro 1 and gyro 2. These datasets include random rotations, pure steady-state conditions, and the combination of both, providing a comprehensive basis to evaluate the accuracy and noise levels of ML-calibrated rotation rates under different test conditions. By incorporating random rotation rates and steady-state periods into the measurement process, we ensure that the highly diverse data used to train and test the models reflect the variability and complexity of real-world conditions. Thereby enhancing the generalizability and reliability of the ML and DL models.

Furthermore, by using two different prototypes of the same MEMS gyroscope design (i.e. gyro 1 and gyro 2), we account for device-to-device variations, which are common in manufacturing processes. This variation is crucial for developing models that are not only accurate but also adaptable to different units, ensuring consistent performance of the models across multiple devices.

### C. Performance Indicator

The use of Mean Squared Error (MSE) and $R^2$ score in evaluating the calibration model is essential for assessing both the accuracy and explanatory power of this regression project [22]. MSE measures the average squared difference between the observed actual outcomes and the predicted values, providing a clear indication of the model's prediction error magnitude. A lower MSE signifies a model that closely aligns with the true rotation rates, which is crucial for precise calibration. Mean Squared Error (MSE) is calculated by:

$$MSE = \frac{1}{n}\sum_{i=1}^{n}(y_i - \hat{y}_i)^2 \quad (1)$$

On the other hand, the $R^2$ score represents the proportion of variance in the dependent variable that is predictable from the independent variables. It provides insight into the model's ability to capture the underlying data patterns. The formula for $R^2$ score is:

$$R^2 = 1 - \frac{\sum_{i=1}^{n}(y_i-\hat{y}_i)^2}{\sum_{i=1}^{n}(y_i-\bar{y}_i)^2} \quad (2)$$

Together, MSE and $R^2$ scores offer a comprehensive evaluation of the model's performance, ensuring that it not only predicts accurately but also explains the variability in the data effectively. This dual assessment is vital for developing a reliable and accurate calibration model for the MEMS gyroscope.

## III. Linear Calibration

As stated in the introduction, the linear fitted scale factor is the traditional method for converting the MEMS gyroscope output signal to a rotation rate, which involves fitting and extracting the constant scale factor in its initial linear region.

### A. Linear Fitted Scale Factor

The linear fitted rotation rate (ω) can be derived from the equation below:

$$\omega = S_{linear} \times V_{sense-in} \quad (3)$$

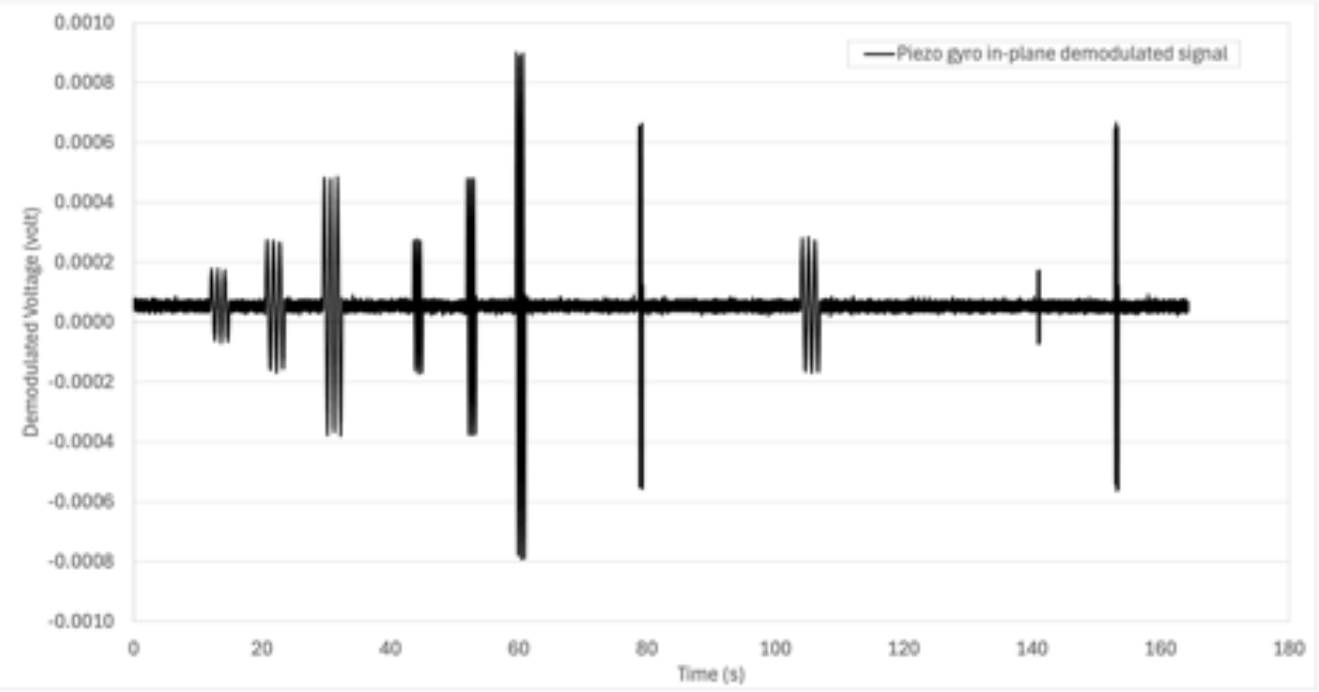

Fig. 4. The in-plane demodulated voltage signal of gyro 1 under controlled rates.

TABLE I

The scale factors for different peaks in Fig. 4, when the rate table changes the input rotation rate under control.

| Peaks | Input Rotation Rate (Degree/Second) | Scale Factor °/(MV·s) | Scale Factor (nA·s/°) |
|---|---|---|---|
| 1st | 20 | 0.161 | 0.623 |
| 2nd | 40 | 0.183 | 0.545 |
| 3rd | 80 | 0.188 | 0.533 |
| 4th | 40 | 0.180 | 0.555 |
| 5th | 80 | 0.188 | 0.533 |
| 6th | 160 | 0.189 | 0.529 |
| 7th | 120 | 0.198 | 0.504 |
| 8th | 40 | 0.177 | 0.566 |
| 9th | 20 | 0.169 | 0.593 |
| 10th | 120 | 0.196 | 0.511 |

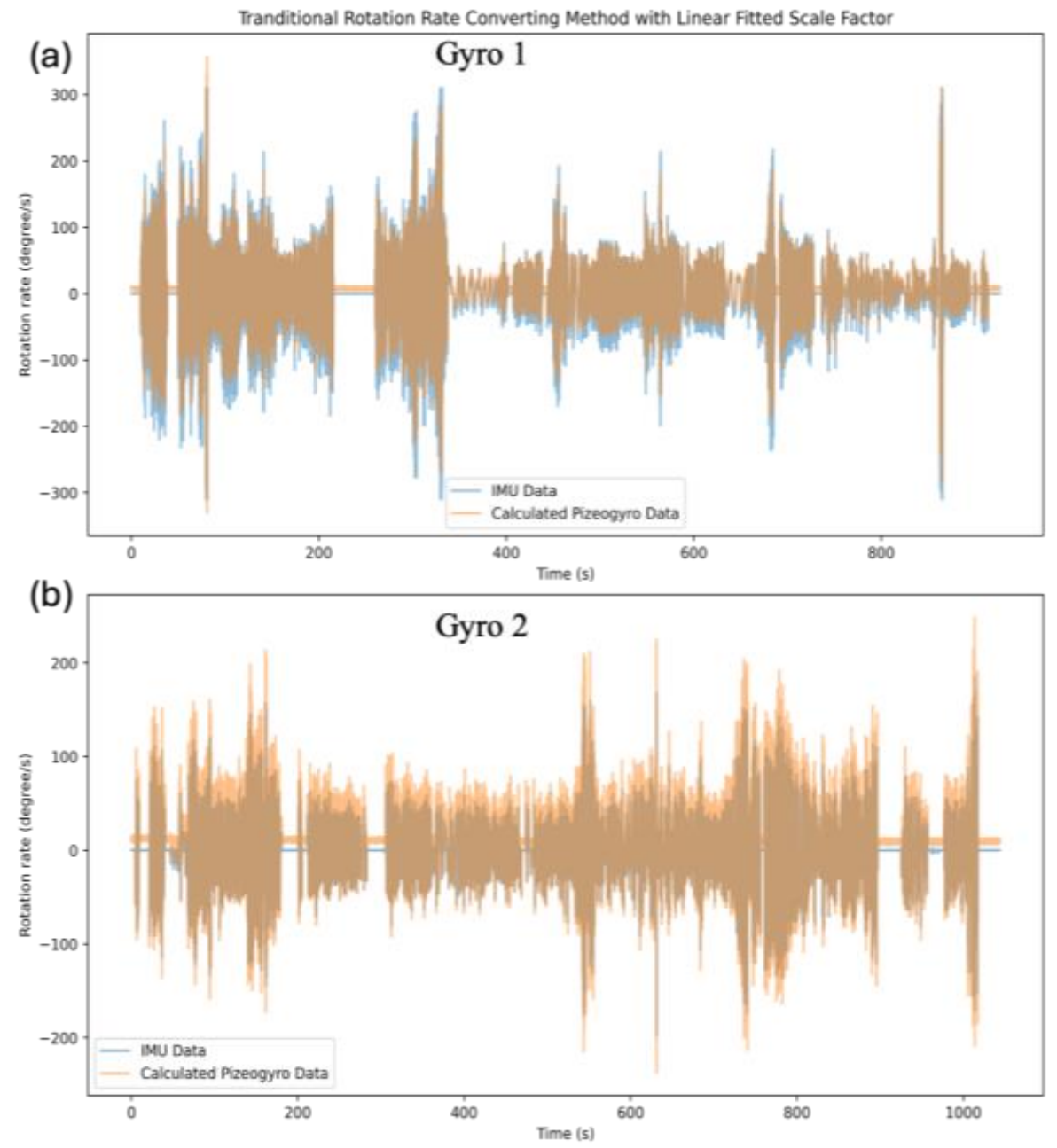

Fig. 5. The linear calibrated rotation rates when testing (a) gyro 1 and (b) gyro 2.

The constant scale factor ($S_{linear}$) is derived from a controlled input rotation rate over the sense mode in-plane signals (demodulated voltage signals, $V_{sense\text{-}in}$) in the first controlled signal response, as shown in Fig. 4. The scale factor is derived to be 0.161 °/(MV·s) from the training gyroscope, gyro 1, as shown in TABLE I.

### B. Linear Fitted Results and Discussion

Using this linear fitted scale factor, 0.161 °/(MV·s), the $R^2$ score of the linear calibrated gyro 1 rotation rate is 0.785. As shown in Fig. 5, the linear calibrated data has a larger difference with the reference IMU readings when predicting the extreme rotation rates including near zero and above 100°/s and negative rotation rate. The same deviations occur and are even worse when using it with gyro 2.

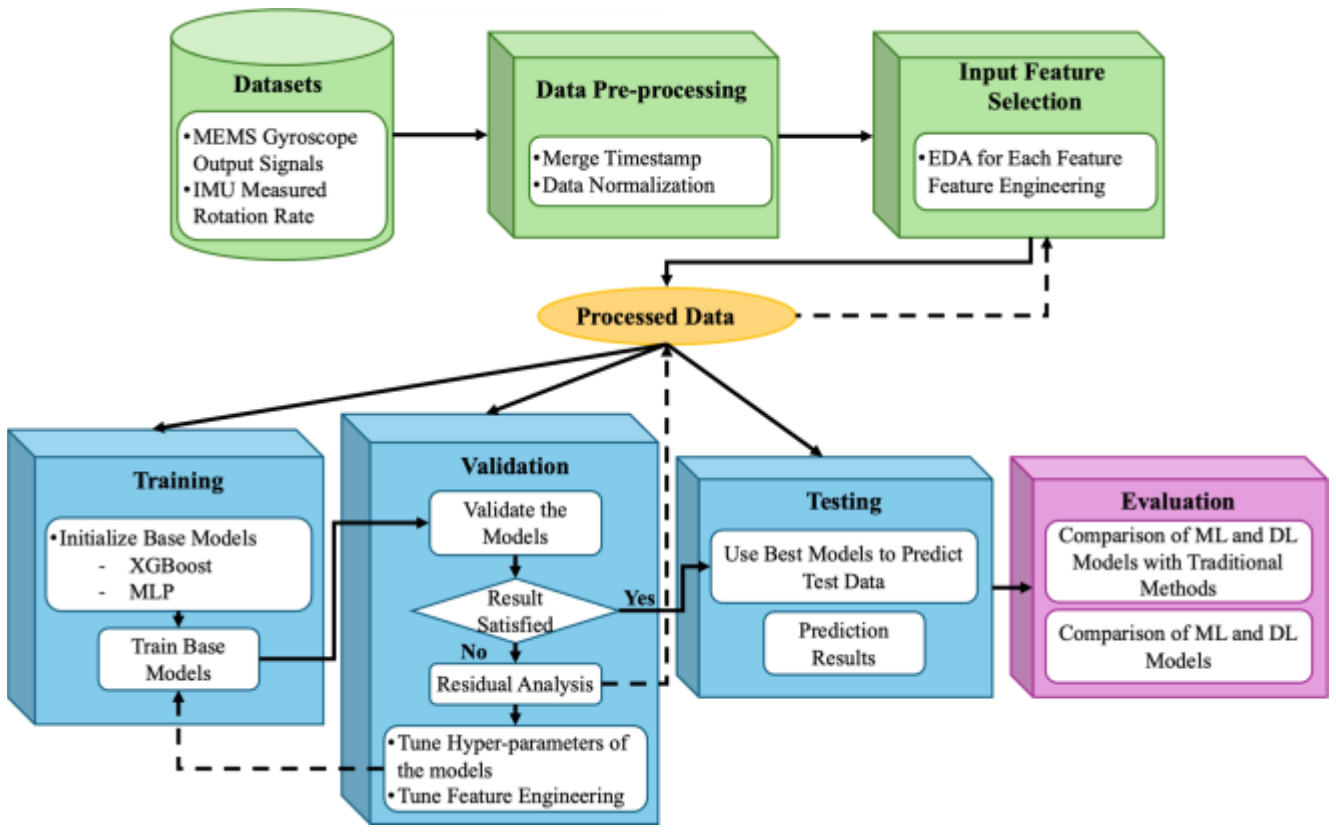

Fig. 6. Flowchart of developing ML models for MEMS gyroscope calibration.

## IV. Machine Learning and Deep Learning Calibrations

Fig. 6 shows the schematic development process of the ML models used to enhance MEMS gyroscope calibration at room temperature. The following sections are also processed in this sequence.

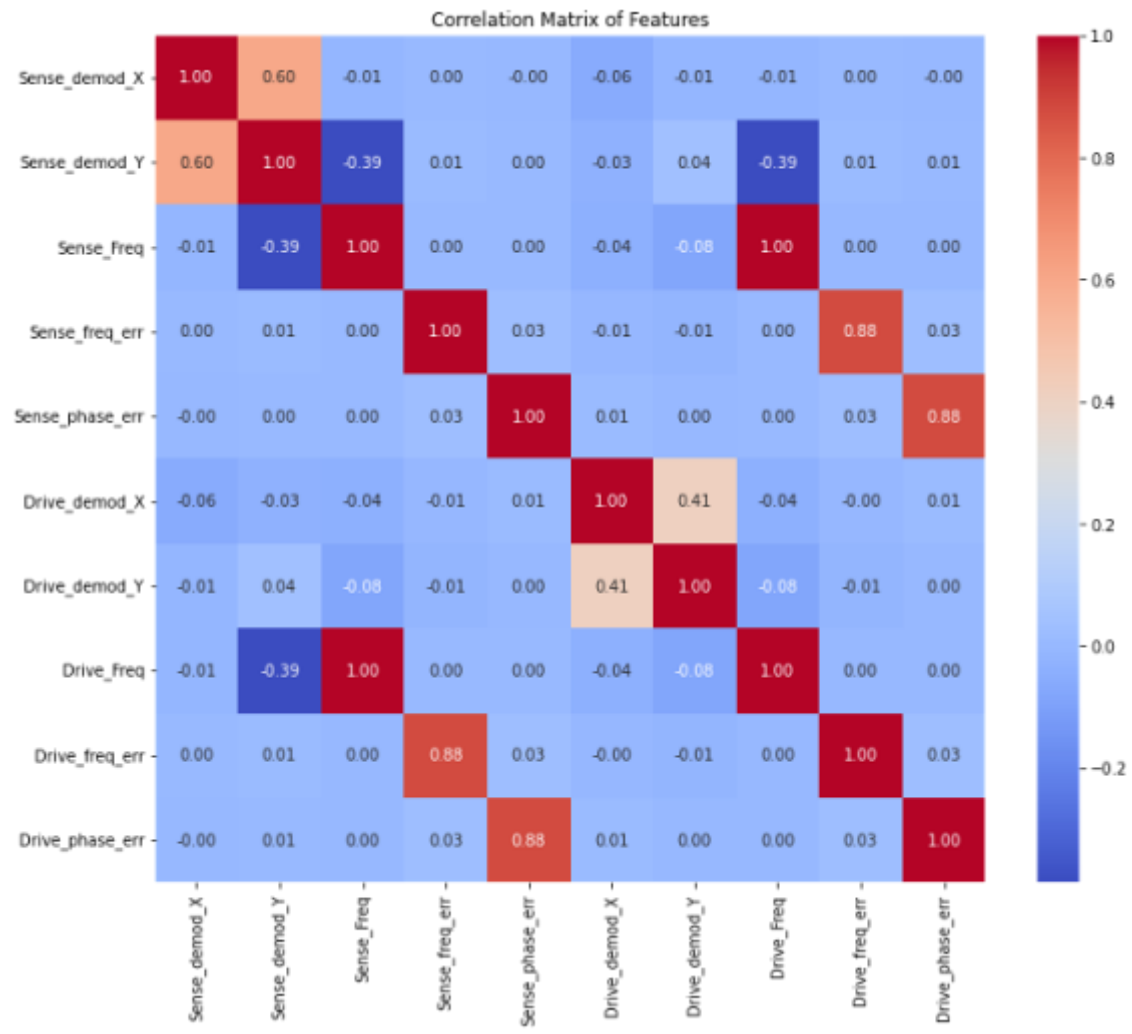

Fig. 7. Feature correlation matrix of the gyroscope output signals (features).

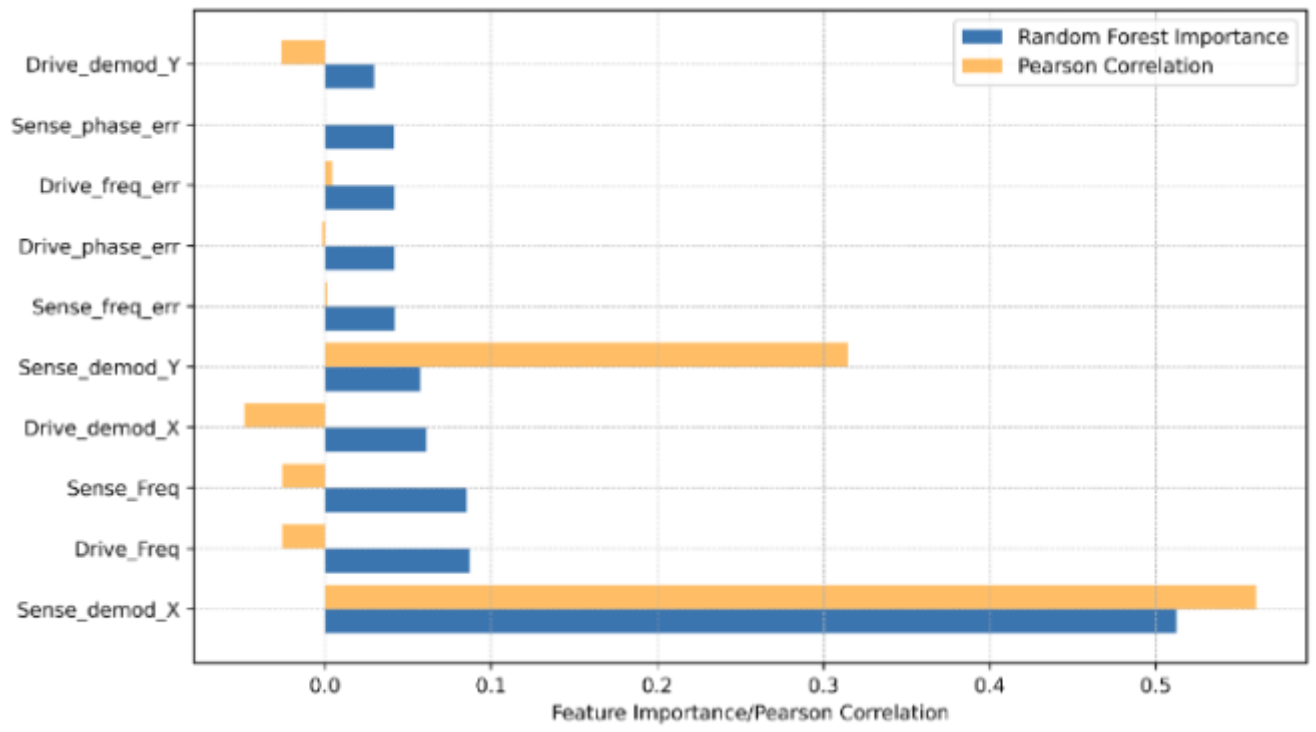

Fig. 8. Gyroscope output signals (features) importance and correlation to the rotation rate.

### A. Feature Analysis

To achieve optimal performance and minimize computational costs, feature selection is necessary for developing ML models. The feature correlation matrix is presented in Fig. 7.

From this matrix, it can be observed that the drive mode frequency and sense mode frequency are nearly identical, and based on the feature importance analysis (Fig. 8), the sense mode frequency is selected. Additionally, the frequency and phase errors of the sense mode are highly correlated with those of the drive mode, exhibiting a correlation value of 0.88 as shown in Fig. 7. Based on feature importance analysis (Fig. 8), the drive mode phase error and the sense mode frequency are included for building the models. The in-plane and quadrature demodulated voltage signals for the sense mode show a moderate correlation, with a value of 0.60, but both of them have higher importance based on Fig. 8. Therefore, these features are also included. Despite these, other features display weak or no correlation with each other, indicating that they can be considered relatively independent. Consequently, the sense mode in-plane signal, sense mode quadrature, sense mode frequency, sense mode frequency error, drive mode in-plane signal, drive mode quadrature, and drive mode phase error are the input features selected for the ML training and testing process.

### B. Feature Engineering

Lagged features are derived features in time series analysis that represent past values of a variable. By shifting the original time series data by a specific number of time steps, these past values can be used as additional input features for predictive models [23, 24]. Since the gyroscope readings are inherently time-dependent, meaning that the current rotation rate is influenced by past rotation rates, lagged features can be efficient in enhancing the performance of predictive regression models by capturing temporal dependencies and providing additional context.

### C. XGBoost Model

With the ability to handle complex, non-linear relationships, the extreme gradient boosting (XGBoost) regression model is renowned for its high accuracy in predicting continuous values [25], like the rotation rate. Additionally, the regularization techniques of XGBoost are highly effective in preventing overfitting and enhancing the model's generalization capabilities. The algorithm is also optimized for speed and scalability, making it suitable for large datasets [25], which are typically encountered in continuous rotation measurements. XGBoost's robustness to outliers and its ensemble learning nature further improve prediction reliability [25]. These features of XGBoost collectively make it an ideal choice for calibrating the MEMS gyroscope. By capturing the complex relationships between the resonator signals and the rotation rate, XGBoost leads to more accurate and reliable calibration and optimization rotation rates.

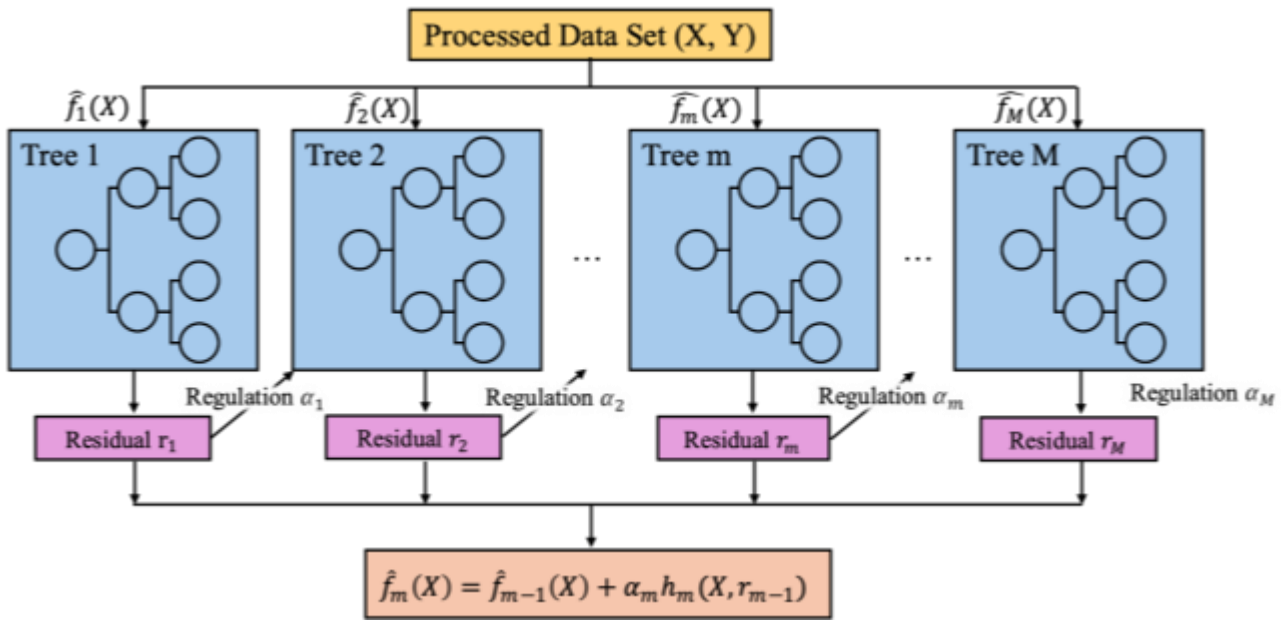

Fig. 9. XGBoost algorithm flowchart (M=50) for gyroscope calibration.

XGBoost is an implementation of gradient boosting that constructs additive models by fitting new models to the residual errors made by prior models. Fig. 9 illustrates the algorism of XGBoost, which constructs additive models by fitting a new tree ($\hat{f}_m(X)$) to the residual errors rm-1 made by prior tree rm-1 made by prior tree ($\hat{f}_{m-1}(X)$) to minimize the objective function:

$$\sum_{m=1}^{M} L\left(Y_m,\ F_{m-1}(X_m) + \alpha h_m(X_m, r_{m-1})\right) \quad (4)$$

where $h_m$ is the regulation term and $\alpha$m is the regulation parameter.

## D. MLP Model

Multilayer Perceptron (MLP) is a fundamental component of the broader field of deep learning. An MLP is a type of artificial neural network (ANN) that consists of multiple layers of nodes, or neurons, with each layer fully connected to the next [26]. As shown in Fig. 10, the basic architecture includes an input layer, one or more hidden layers, and an output layer. Each neuron in a layer receives inputs, applies a linear combination of weights and biases, and passes the result through a nonlinear activation function. In this gyroscope calibration process, the Exponential

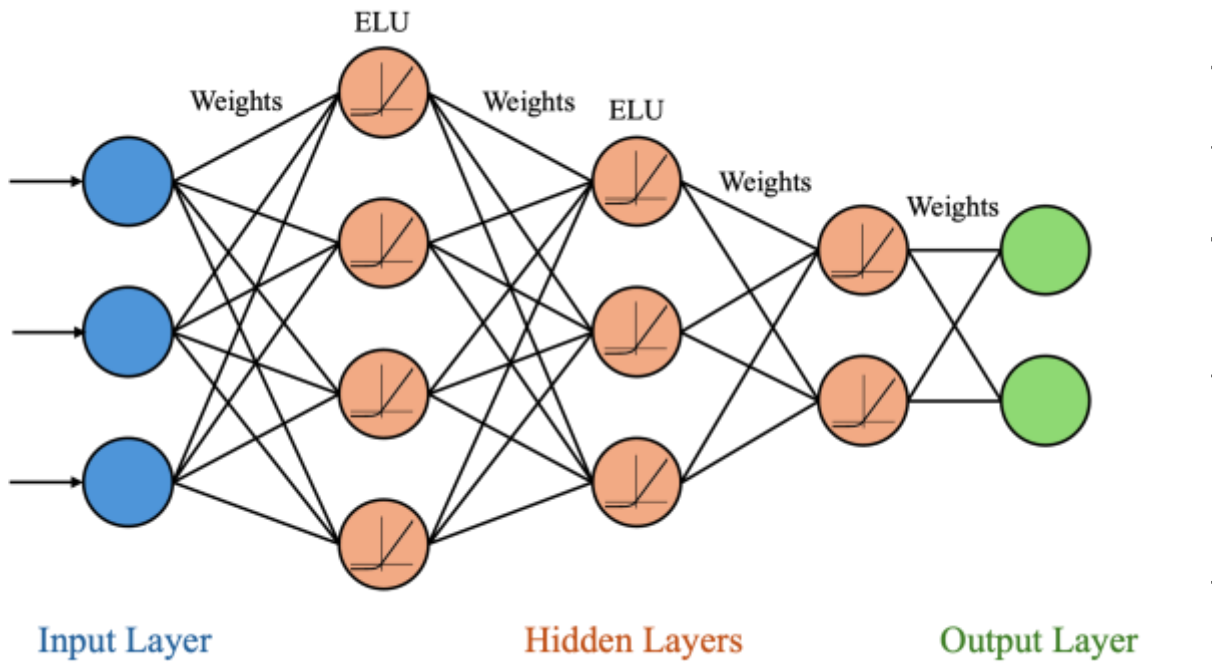

Fig. 10. MLP algorithm schematic flowchart (Not represent this application. 3 hidden layers for gyroscope calibration).

Linear Unit (ELU) is used as the activation function due to its smooth, non-linear nature [27], which enables the model to capture complex, non-linear relationships in the continuous MEMS gyroscope data more effectively. Additionally, MLPs can handle missing values and noise in the data [26], which are common challenges in the gyroscope calibration process. The robustness of MLPs to such imperfections ensures that the calibration model remains reliable and accurate even in less ideal conditions. This capability is also valuable for enhancing the performance of the MEMS gyroscope, ensuring precise calibration and optimized outcomes despite potential data quality issues.

## E. XGBoost & MLP Calibrated Results and Discussions

As stated in II.B, both XGBoost and MLP models were trained with gyro 1 using a highly diverse dataset. Following the validation process, the tuned XGBoost regression model and the MLP model were used to test the datasets collected from gyro 1 and gyro 2. The XGBoost and MLP calibrated rotation rates are presented in Fig. 11, and the performance metrics for each model across different test gyroscopes are summarized in TABLE II. A comparison of Fig. 11 with Fig. 5 and TABLE II reveals that the MSE significantly decreased and the $R^2$ score substantially increased with the same test gyroscope when using XGBoost and MLP calibration methods. This indicates that both XGBoost and MLP models markedly improve the calibration performance of the MEMS gyroscopes across both test gyroscopes.

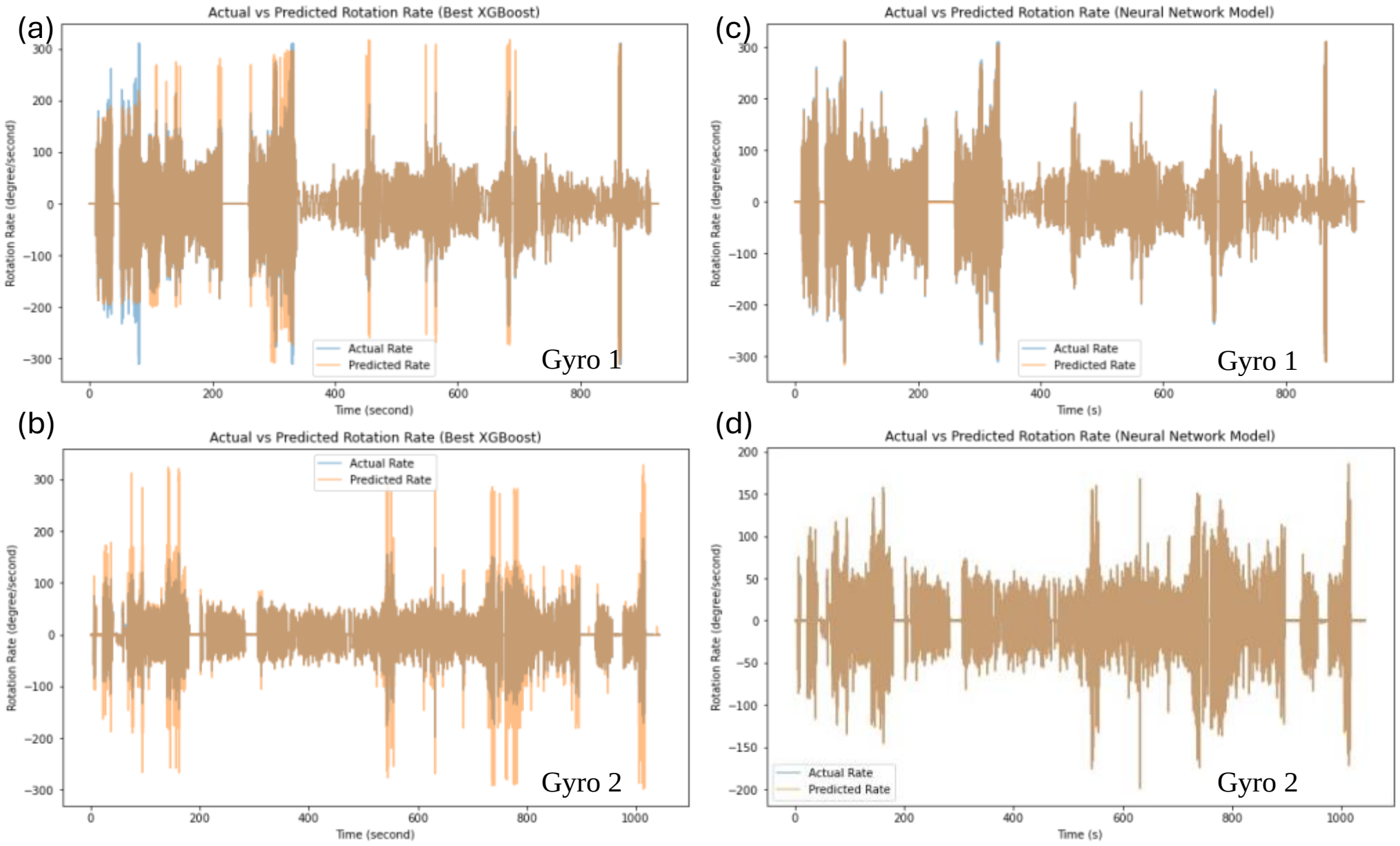

Fig. 11. The XGBoost calibrated rotation rates when testing (a) gyro 1 and (b) gyro 2. The MLP calibrated rotation rates when testing (c) gyro 1 and (d) gyro 2.

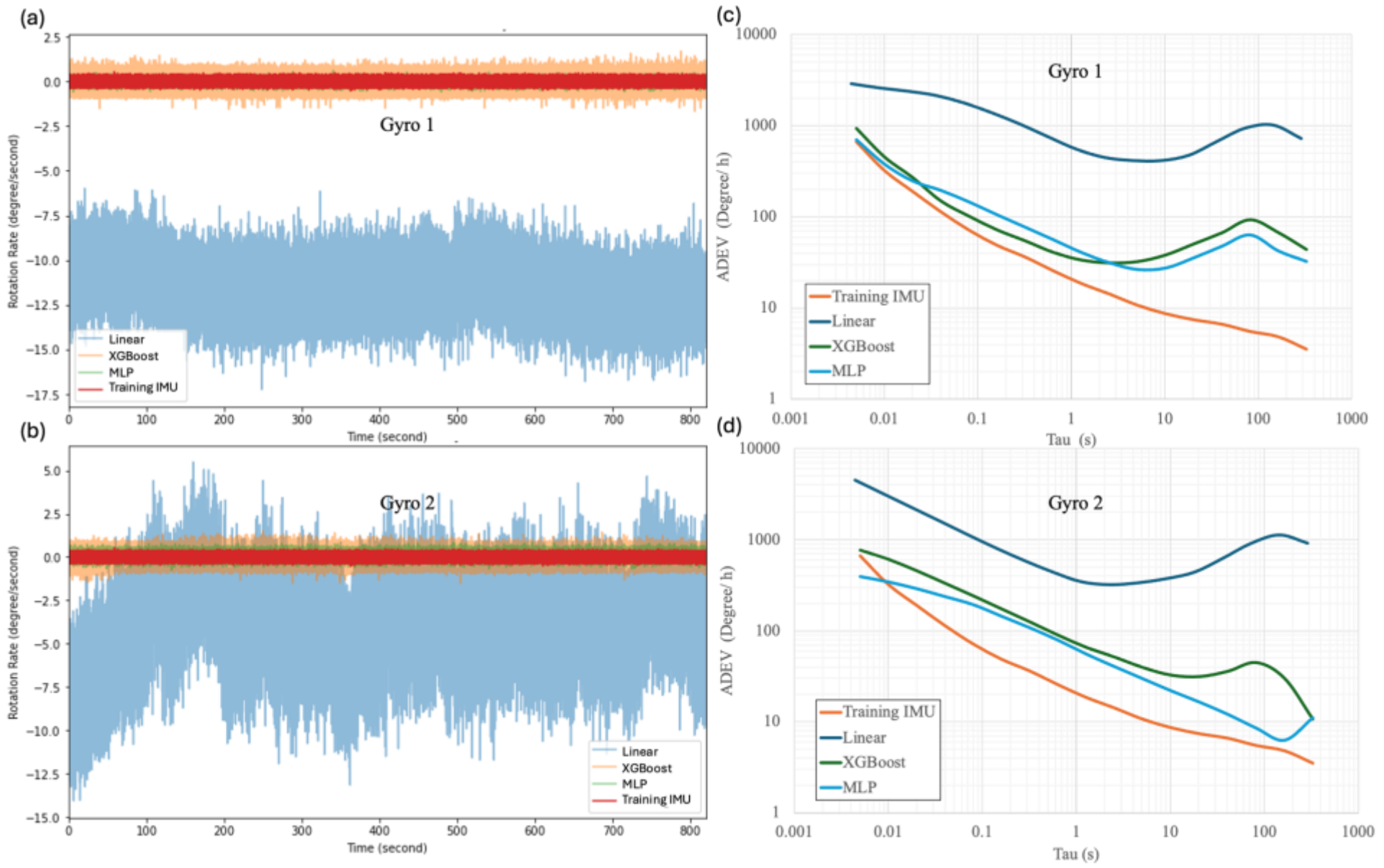

Fig. 12. (a) and (b) are the comparisons of reference IMU rotation rate readings of training data with linear, XGBoost and MLP (mostly overlapped with training IMU) calibrated rotation rates at steady state with (a) gyro 1 and (b) gyro 2. (c) and (d) are the ADEV plots of the linear, XGBoost and MLP calibrated rotation rates compared with the reference IMU with (a) gyro 1 and (b) gyro 2.

When the same gyroscope (gyro 1) is used for both training and testing processes, both XGBoost and MLP models can well calibrate the rotation rate. In Fig. 11, the XGBoost calibrated rotation rates show a strong alignment with the reference IMU rotation rates, and the MLP calibrated rates exhibit an even tighter fit, demonstrating superior accuracy. This is corroborated by TABLE II, which indicates that the MLP method achieves the lowest MSE (0.5014) and the highest $R^2$ score (0.9998), while the XGBoost method also performs well, with an MSE of 115.4446 and an $R^2$ score of 0.9688. In comparison, the traditional method with a linear fitted scale factor has a higher MSE and a lower $R^2$ score, highlighting the advantages of using XGBoost and MLP techniques for the calibration of the same gyroscope.

When different gyroscopes are used for training (gyro 1) and testing (gyro 2), the MLP model experiences an almost negligible decline in performance compared with the condition using the same gyroscope for training and testing, while the XGBoost model experiences a more notable decrease. Fig. 11 also shows significant deviations and variability in the predicted rotation rates, with notable outliers present in XGBoost model calibrated rotation rates. Despite this, the MLP model maintains superior performance compared to the XGBoost model, as indicated by the lower MSE (0.3557) and higher $R^2$ score (0.9997) in TABLE II. The XGBoost method, while performing reasonably well with an MSE of 202.5524 and an $R^2$ score of 0.8397, does not match the accuracy and robustness of the MLP method. However, the traditional calibration methods still show higher MSEs and lower $R^2$ scores, underscoring the benefits of adopting advanced XGBoost and MLP techniques for gyroscope calibration, particularly in conditions involving different devices.

Overall, the combined analysis of Fig. 11 with Fig. 5 and TABLE II demonstrates that MLP methods consistently outperform XGBoost and traditional linear fitting methods in calibrating the MEMS gyroscope across two test gyroscopes, gyro1 and gyro 2. The superior performance of MLP methods is evident in their lowest MSEs, highest $R^2$ scores, and better handling of variability and generalization across different gyroscopes. These findings highlight the value of incorporating advanced XGBoost and MLP techniques in improving the accuracy and reliability of gyroscope calibration.

Moreover, the denoising results and the ADEV plots of the XGBoost and MLP calibrated rotation rates of gyro 1 and gyro 2 under salient conditions are shown in Fig. 12. Fig. 12a and 12b compare the training reference IMU readings with linear fitted, XGBoost and MLP calibrated rotation rates at steady state. For gyro 1, Fig. 12a shows that the traditional calibration method exhibits significant noise and variability, while both XGBoost and MLP methods significantly reduce this noise and variability, with the MLP method showing the most stable and consistent readings. For gyro 2, Fig. 12b shows that the traditional method again shows considerable noise and variability, while the XGBoost method provides improved stability but still has some fluctuations. The MLP method, however, consistently offers the most stable and accurate rotation rate readings, demonstrating superior performance in both scenarios.

Fig. 12c and 12d present the ADEV plots of traditional, XGBoost, and MLP calibrated rotation rates compared with the training reference IMU readings when testing gyro 1 and gyro 2. For gyro 1, the ADEV plot of linear calibrated rotation rates (Fig. 12c) shows a high deviation, indicating significant noise over various time intervals. The XGBoost method reduces ADEV, indicating better stability, while the MLP method achieves the lowest ADEV values across almost all tau (τ) values, signifying superior noise reduction and stability. When gyro 2 is tested (Fig. 12d), the XGBoost method demonstrates improved stability with lower ML-enhanced ADEV values, but the MLP method consistently achieves the lowest ADEV values across a wide range of τ values. This underscores the MLP method's robust capability in reducing noise and ensuring stable measurements, even with different devices. TABLE III summarizes the ARW and the BI of the calibrated rotation rate by traditional linear fitted scale factor method, XGBoost, and MLP model enhanced methods. The reduction in ADEV following the application of ML models is due to the ability of models to identify patterns in the subtle variations of gyroscope output signal datasets between the steady state and low rotation rates. This improvement is achieved by leveraging multiple gyroscope output signals. Consequently, the integration of gyroscopes with ML models significantly reduces noise levels and enhances the detection of lower rotation rates compared to the traditional approach of using only a linear fitted scale factor and a single resonator output signal (i.e., sense mode in-phase amplitude). This is evidenced by the comparison of the calibrated rotation rate around 950s in Fig. 11b and 11d with Fig. 5b, where the reference IMU indicates rotation rates of approximately 3°/s. In Fig. 5b, the rotation rate is obscured by noise after linear fitted scale factor calibration, whereas the ML-calibrated rotation rate is clearly around 3°/s, aligning with the IMU reference readings shown in Fig. 11b and 11d. As illustrated in TABLE III, the ML-calibrated ARW values are close to the Brownian noise level of the gyroscopes, which is approximately 0.44 °/√h in this study. Due to the inherent fabrication variances, the calibrated ARW of the gyroscope using both training and testing is further reduced and approaches the Brownian noise level.

TABLE III

The performance of traditional (linear fitted), XGBoost and MLP calibrated rotation rates at steady state.

| | | Same Gyroscope (gyro 1) | | | Different Gyroscope (gyro 2) | | |
|---|---|---|---|---|---|---|---|
| Method | Training IMU | Traditional | XGBoost | MLP | Traditional | XGBoost | MLP |
| ARW (°/√h ) | 0.3521 | 9.8148 | 0.6369 | 0.5851 | 6.1613 | 1.2398 | 0.9095 |
| BI (°/h ) | 3.5376 | 414.5842 | 31.4438 | 26.5193 | 322.1151 | 10.7516 | 6.3060 |

As stated above, when compared with the XGBoost model, the MLP model outperforms regarding its outstanding stability, accuracy, and adaptability to different devices, but it needs much more computational costs, whereas the training and testing time of the XGBoost model for the same dataset is around thirty times lower than the MLP model. Since the testing time can potentially impact the bandwidth of the system, particularly if the data processing rate (processed data points per second) is lower than the bandwidth of gyroscopes. Extended testing times may hinder the overall system performance. In Fig. 13, the residual analysis of the XGBoost-calibrated rotation rate of gyro 1 and gyro 2 is produced to better understand the source of the XGBoost model imperfection, which enables the selection between XGBoost and MLP models for more suitable application areas.

In Fig. 13a and 13b, the test device is gyro 1, i.e. the same gyroscope is used during both training and testing; the residuals are concentrated around the zero line, with some deviations at higher magnitudes, indicating good model performance with minimal errors. When testing gyro 2, Fig. 13c and 13d show notable deviations and a narrow peak at zero with significant outliers, indicating variability and inconsistency in model performance. Overall, the presence of significant residuals at extreme rotation rate values and heteroscedasticity affect the accuracy of the XGBoost model's predictions for both test gyroscopes.

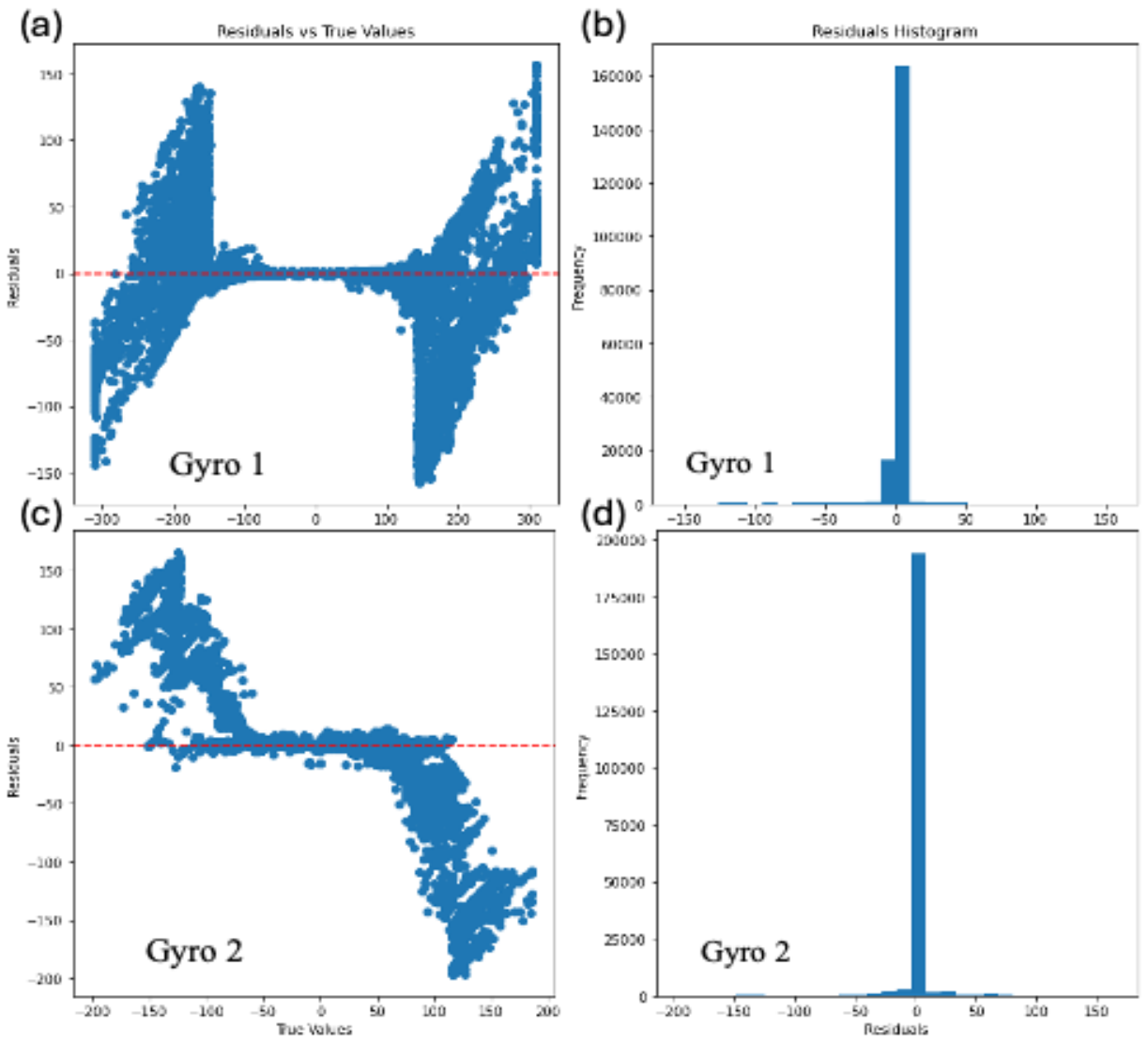

Fig. 13. The residual analysis results of XGBoost-calibrated rotation rate, compared with the reference IMU readings, in three different scenarios. The left column is the residuals vs true values (IMU readings) plot. The right column is the residual histogram.

In summary, the XGBoost model's lower computational cost and moderate accuracy make it suitable for the following gyroscope application areas. In consumer electronics, such as smartphones and wearable devices, the XGBoost model's efficiency and acceptable performance are ideal given the limited processing power and battery life constraints. Health monitoring applications also benefit from the XGBoost model's cost-efficiency and sufficient accuracy for tracking vital signs and other metrics in non-critical scenarios. Additionally, environmental monitoring, which involves deploying numerous sensors across wide areas, requires cost-effective solutions. The XGBoost model's lower computational demands align well with the limited processing capabilities of remote sensors, providing adequate precision for tasks like air quality monitoring and basic weather prediction.

Meanwhile, the MLP model's superior stability, accuracy, and adaptability to different environmental conditions make it well-suited for high-stakes and complex applications. In aerospace and defense, the MLP model's precision and robustness are crucial for ensuring reliable performance in diverse and unpredictable environments. Autonomous vehicles also benefit from the MLP model's high accuracy and ability to handle varying conditions, which are essential for safe and efficient navigation. Moreover, industrial automation applications demand precision and consistency, which the MLP model provides by reducing noise and variability in sensor measurements. Although the MLP model requires higher computational resources, the critical nature of these applications justifies the investment in advanced processing capabilities.

## V. Conclusion

This study demonstrates the significant advances that ML and DL could bring to the calibration of MEMS gyroscopes. Traditional calibration methods, which involve fitting a linear scale factor for each gyroscope, fall short of capturing the intricate, non-linear relationships between the gyroscope output and the rotation rate. These methods also demand extensive manual labor and precise control over each gyroscope, making them time-consuming and costly.

Our research shows that both ML and DL methods significantly enhance the accuracy and reliability of MEMS gyroscopes by effectively modeling complex patterns and relationships. DL techniques consistently achieved the lowest MSEs and highest $R^2$ scores, outperforming both ML and traditional methods. Despite the higher computational costs associated with DL models, their superior performance justifies their use in high-stakes applications such as aerospace, defense, and autonomous vehicles, where precision and robustness are critical. Furthermore, ML models, while not as accurate as DL models, offer a good balance between computational cost and performance. Their efficiency makes them suitable for applications with limited processing power and battery life, such as consumer electronics, health monitoring, and environmental monitoring.

Overall, the integration of ML and DL in MEMS gyroscope calibration provides a robust solution to existing challenges, enhancing performance, reducing noise, and ensuring stability across various devices and applications. This study underscores the transformative potential of advanced machine learning techniques in improving the calibration and functionality of MEMS gyroscopes, paving the way for more reliable and efficient devices.